# STREAMING PUNCTUATION FOR LONG-FORM DICTATION WITH TRANSFORMERS

Piyush Behre, Sharman Tan, Padma Varadharajan and Shuangyu Chang

Microsoft Corporation, USA

***ABSTRACT***

*While speech recognition Word Error Rate (WER) has reached human parity for English, long-form dictation scenarios still suffer from segmentation and punctuation problems resulting from irregular pausing patterns or slow speakers. Transformer sequence tagging models are effective at capturing long bi-directional context, which is crucial for automatic punctuation. Automatic Speech Recognition (ASR) production systems, however, are constrained by real-time requirements, making it hard to incorporate the right context when making punctuation decisions. In this paper, we propose a streaming approach for punctuation or re-punctuation of ASR output using dynamic decoding windows and measure its impact on punctuation and segmentation accuracy across scenarios. The new system tackles over-segmentation issues, improving segmentation $F_{0.5}$-score by 13.9%. Streaming punctuation achieves an average BLEU-score improvement of 0.66 for the downstream task of Machine Translation (MT).*

***KEYWORDS***

*Automatic punctuation, automatic speech recognition, re-punctuation, speech segmentation.*

## 1. INTRODUCTION

Our hybrid Automatic Speech Recognition (ASR) generates punctuation with two systems working together. First, the Decoder generates text segments and passes them to the Display Post Processor (DPP). The DPP system then applies punctuation to these segments.

This setup works well for single-shot use cases like voice assistant or voice search but fails for long-form dictation. A dictation session typically comprises many spoken-form text segments generated by the decoder. Decoder features such as speaker pause duration determine the segment boundaries. The punctuation model in DPP then punctuates each of those segments. Without cross-segment look-ahead or the ability to correct previously finalized results, the punctuation model functions within the boundaries of each provided text segment. Consequently, punctuation model performance is highly dependent on the quality of text segments generated by the decoder.

Our past investments have focused on both systems independently - (1) improving decoder segmentation using look-ahead-based acoustic-linguistic features and (2) using neural network architectures to punctuate in DPP. As measured by punctuation-$F_1$ scores, these investments have improved our punctuation quality. Yet, over-segmentation for slow speakers or irregular pauses is still prominent.

In streaming punctuation, we explore a system that discards decoder segmentation, shifting punctuation decision-making towards a powerful long-context Transformer-based punctuation model. Instead of segments, this system emits well-formed sentences, which is much more desirable for downstream tasks like Translation of Speech Recognition output. The proposed architecture also satisfies real-time latency constraints for ASR.

Many works have demonstrated that leveraging prosodic features and audio inputs can improve punctuation quality [1, 2, 3]. However, as we show in our experiments, misleading pauses may undermine punctuation quality and encourage overly aggressive punctuation. It is especially true in scenarios like dictation, where users pause often and unintentionally. Our work shows that text-only streaming punctuation is robust to over-segmentation from irregular pauses and slow speakers.

We make the following key contributions: (1) we introduce a novel streaming punctuation approach to punctuate and re-punctuate ASR outputs, as described in section 3, (2) we demonstrate streaming punctuation's robustness to model architecture choices through experiments described in section 5, and (3) we achieve not only gains in punctuation quality but also significant downstream BLEU score gains on Machine Translation for a set of languages, as demonstrated in section 6.

## 2. RELATED WORK

Approaches to punctuation restoration have evolved to capture the surrounding context better. Early sequence labeling approaches for punctuation restoration used n-grams to capture context [4]. However, this simple approach becomes unscalable as n grows large, limiting the amount of context used in punctuation prediction.

Classical machine learning approaches such as conditional random fields (CRFs) [5, 6, 7], maximum entropy models [8], and hidden Markov models (HMMs) [9] model complex features by leveraging manual feature engineering. This manual process is cumbersome, and the quality of these features limits the effectiveness of these classical approaches.

Neural approaches mostly displaced manual feature engineering, opting instead to learn more complex features through deep neural models. Recurrent neural networks (RNNs), specifically Gated Recurrent Units (GRUs) and bidirectional Long Short-Term Memory (LSTM) networks, have advanced natural language processing (NLP) and punctuation restoration by specifically modeling long-term dependencies in the text [10, 11, 12, 13, 14]. Prior works have also successfully used LSTMs with CRF layers [15, 16]. Most recently, using Transformers [17] and especially pre-trained embeddings from models such as Bidirectional Encoder Representations from Transformers (BERT) [18] has significantly advanced quality across natural language processing (NLP) tasks. By leveraging attention mechanisms and more complex model architectures, Transformers better capture bidirectional long-range text dependencies for punctuation restoration [19, 20, 21, 22, 23, 24].

## 3. PROPOSED METHOD

### 3.1. Punctuation Model

We frame punctuation prediction as a sequence tagging problem. Figure 1 illustrates the end-to-end punctuation tagging and application process. We first tokenize the input segment as byte-pair encoding (BPE) tokens and pass this through a transformer encoder. Next, a punctuation token classification head, consisting of a dropout layer and a fully connected layer, generates token-level punctuation tags. Finally, we convert the token-level tags to word-level and generate punctuated text by appending each specified punctuation symbol to the corresponding word in the input segment.

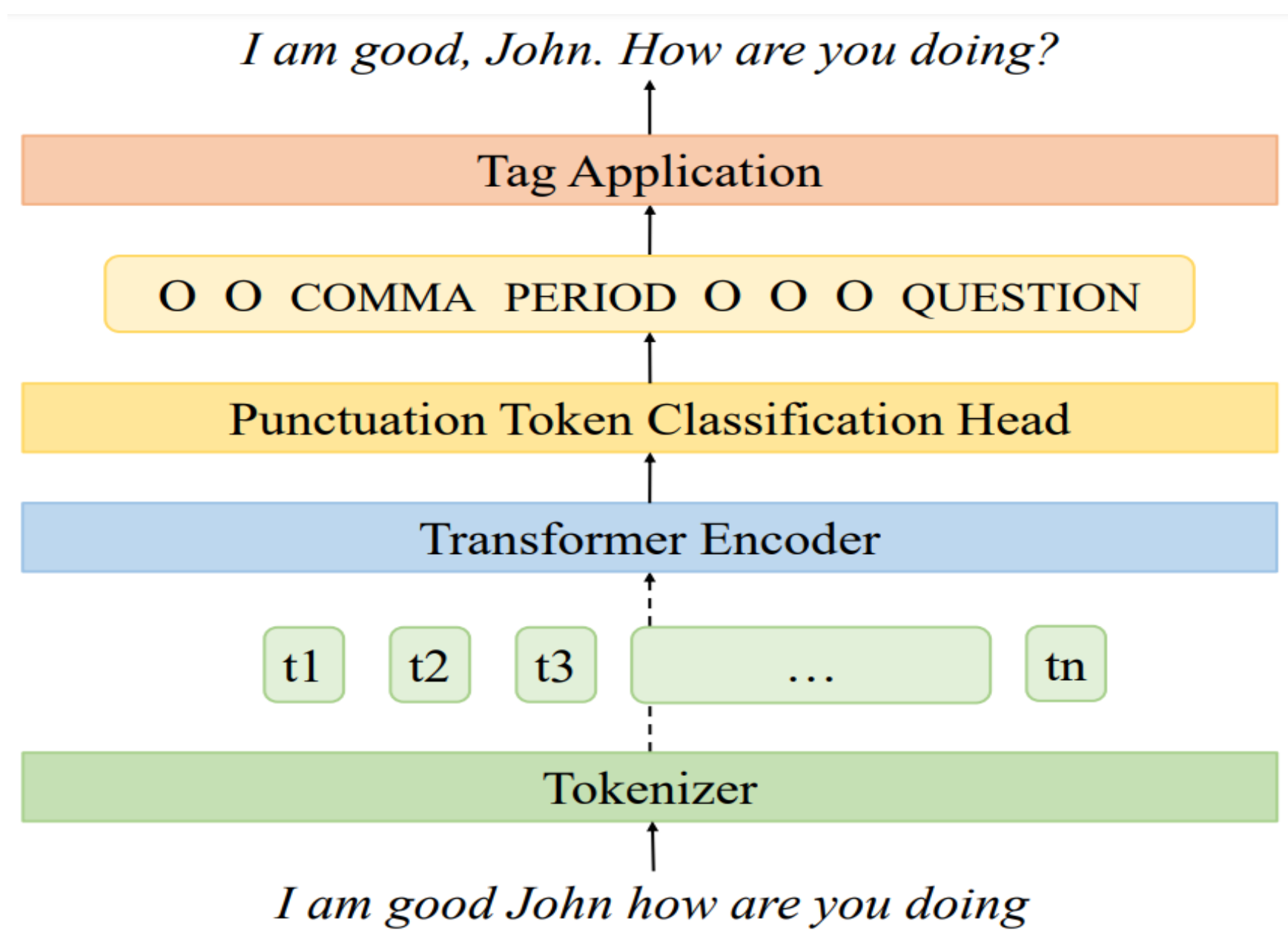

Figure 1. Punctuation tagging model using transformer encoder

## 3.2. Streaming Decoder for Punctuation

Hybrid ASR systems often define segmentation boundaries using silence thresholds. However, for human2machine scenarios like dictation, pauses do not necessarily indicate ideal segmentation boundaries for the ASR system. In our experience, users pause at unpredictable moments as they stop to think. All A1-4 segments in Table 1 are possible; each is a valid sentence with correct punctuation. Even with a punctuation model, if A4 is the user's intended sentence, all A1-3 would be incorrect. For dictation users, this system would produce a lot of over-segmentation. To solve this issue, we must incorporate the right context across segment boundaries.

Table 1. Examples of possible segments generated by ASR

| **Id** | **Segment** |
|---|---|
| A1 | It can happen. |
| A2 | It can happen in New York. |
| A3 | It can happen in New York City. |
| A4 | It can happen in New York City, right? |

Our solution is a streaming punctuation system. The key is to emit complete sentences only after detecting the beginning of a new sentence. At each step, we punctuate text within a dynamic decoding window. This window consists of a buffer for which the system hasn't yet detected a sentence boundary and the new incoming segment. When at least one sentence boundary is detected within the dynamic decoding window, we emit all complete sentences and reserve any remaining text as the new buffer. This process is illustrated in Figure 2.

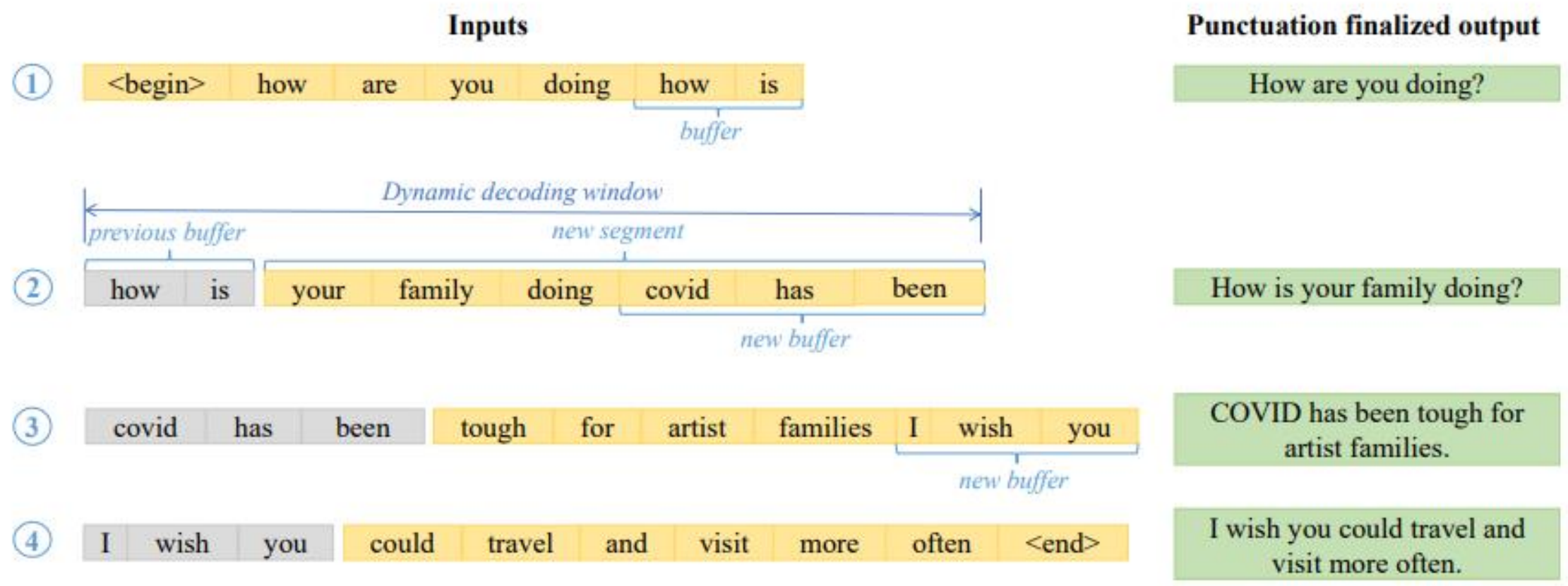

Figure 2. Dynamic decoding window for streaming punctuation

This strategy discards the original decoder boundary and decides the sentence boundary purely on linguistic features. A powerful transformer model that captures the long context well is ideal for this strategy, as dynamic windows ensure that we incorporate enough left and right context before finalizing punctuation. Our approach also meets real-time requirements for ASR without incurring additional user-perceived latency, owing to the continual generation of hypothesis buffers within the same latency constraints. An improvement to this system would be to use a prosody-aware punctuation model that captures both acoustic and linguistic features. That would be a way to re-capture the acoustic cues that we lose by discarding the original segments. However, prosody-aware punctuation models may cause regressions in scenarios such as dictation.

# 4. DATA PROCESSING PIPELINE

## 4.1. Datasets

We use public datasets from various domains to ensure a good mix of conversational and written-form data. Table 2 shows the word count distributions by percentage among the sets.

**OpenWebText** [25]: This dataset consists of web content from URLs shared on Reddit with at least three upvotes.

**Stack Exchange**: This dataset consists of user-contributed content on the Stack Exchange network.

**OpenSubtitles2016** [26]: This dataset consists of movie and TV subtitles.

**Multimodal Aligned Earnings Conference (MAEC)** [27]: This dataset consists of transcribed earnings calls based on S&P 1500 companies.

**National Public Radio (NPR) Podcast**: This dataset consists of transcribed NPR Podcast episodes.

Table 2. Data distribution by number of words per dataset

| Dataset | Distribution |
|---|---|
| OpenWebText | 52.8% |
| Stack Exchange | 31.5% |
| OpenSubtitles2016 | 7.6% |
| MAEC | 6.7% |
| NPR Podcast | 1.4% |

### 4.2. Data Processing

As described in Section 3.1, the transformer sequence tagging model takes spoken-form unpunctuated text as input and outputs a sequence of tags signifying the punctuation to append to the corresponding input word. All datasets consist of punctuated written-form paragraphs, and we process them to generate spoken-form input text and output tag sequences for training.

To preserve the original context, we keep the original paragraph breaks in the datasets and use each paragraph as a training row. We first clean and filter the sets, removing symbols apart from alphanumeric, punctuation, and necessary mid-word symbols such as hyphens. To generate spoken-form unpunctuated data, we strip off all punctuation from the written-form paragraphs and use a Weighted Finite-State Transducers (WFST) based text normalization system to generate spoken-form paragraphs. During text normalization, we preserve alignments between each written-form word and its spoken form. We then use these alignments and the original punctuated display text to generate ground truth punctuation tags corresponding to the spoken-form text.

We set aside 10% or at most 50k paragraphs from each set for validation and use the rest for training.

### 4.3. Tag Classes

We define four tag categories: comma, period, question mark, and ‘O’ for no punctuation. Each punctuation tag represents the punctuation that appears appended to the corresponding text token. When we convert input word sequences into BPE sequences, we attach the tags only to the last BPE token for each word. The rest of the tokens are tagged with ‘O’.

## 5. EXPERIMENTS

### 5.1. Test Sets

We evaluate our punctuation model performance across various scenarios using private and public test sets. Each set contains long-form audio and corresponding written-form transcriptions with number formatting, capitalization, and punctuation. Starting from audio rather than text is critical to highlight the challenges associated with irregular pauses or slow speakers. This prohibits us from using the text-only International Conference on Spoken Language Translation (IWSLT) 2011 TED Talks corpus, typically used for reporting punctuation model performance.

**Dictation (Dict-100)**: This internal set consists of 100 sessions of long-form dictation ASR outputs and corresponding human transcriptions. On average, each session is 180 seconds long.

**MAEC**: 10 hours of test data taken from the MAEC corpus, containing transcribed earnings calls.

**European Parliament (EP-100)**: This dataset contains 100 English sessions scraped from European Parliament Plenary videos. This dataset already contains English transcriptions, and human annotators provided corresponding translations into seven other languages.

**NPR Podcast (NPR-76)**: 20 hours of test data from transcribed NPR Podcast episodes.

### 5.2. Experimental Setup

Our baseline system primarily uses Voice Activity Detection (VAD) based segmentation with a silence-based timeout threshold of 500ms. When VAD doesn't trigger, the system applies a segmentation at 40 seconds. The streaming punctuation system receives the input from the baseline system but can delay finalizing punctuation decisions until it detects the beginning of a new sentence.

We hypothesize that streaming punctuation outperforms the baseline system. We test our hypothesis on LSTM and transformer punctuation tagging models. For the LSTM tagging model, we trained a 1-layer LSTM with 512-dimension word embeddings and 1024 hidden units. We used a look-ahead of 4 words, providing limited right context for better punctuation decisions. For the transformer tagging model, we trained a 12-layer transformer with 16 attention heads, 1024-dimension word embeddings, 4096-dimension fully connected layers, and 8-dimension layers projecting from the transformer encoder to the decoder that maps to the tag classes.

Both models are trained with 32k BPE units. We limited training paragraph lengths to 250 BPE tokens and trimmed paragraphs to the last full sentence. All models are trained to convergence.

## 6. Results and Discussion

We compare the results of baseline (BL) and streaming (ST) punctuation systems on (1) the LSTM tagging model and (2) the Transformer tagging model. As expected, Transformers outperform LSTMs for this task. Here we test our hypothesis for both model types to establish the effectiveness and robustness of our proposed system. For LSTM tagging models, BL-LSTM refers to the baseline system, and ST-LSTM refers to the streaming punctuation system. Similarly, for Transformer tagging models, BL-Transformer refers to the baseline system, and ST-Transformer refers to the streaming punctuation system.

### 6.1. Punctuation and Segmentation Accuracy

We measure and report punctuation accuracy with word-level precision (P), recall (R), and $F_1$-score. Table 3 summarizes punctuation metrics measured and aggregated over three punctuation categories: period, question mark, and comma.

Our customers consistently prefer higher precision (system only acting when confident) over higher recall (system punctuating generously). Punctuation-$F_1$ does not fully capture this preference. Customers also place higher importance on correctly detecting sentence boundaries over commas. We, therefore, propose segmentation-$F_{0.5}$ as a primary metric for this and future sentence segmentation work. The segmentation metric ignores commas and treats periods and question marks interchangeably, thus only measuring the quality of sentence boundaries. Table 4 summarizes segmentation metrics.

Although our target scenario was long-form dictation (human2machine), we found this technique equally beneficial for conversational (human2human) and broadcast (human2group) scenarios, establishing its robustness. On average, the ST-Transformer system has a segmentation-$F_{0.5}$ gain of 13.9% and a punctuation-$F_1$ gain of 4.3% over the BL-Transformer system. Similarly, the ST-LSTM system has a Segmentation-$F_{0.5}$ improvement of 12.2% and a Punctuation-$F_1$ improvement of 2.1% over the BL-LSTM system. These results support our hypothesis.

## 6.2. Downstream Task: Machine Translation

We measure the impact of segmentation and punctuation improvements on the downstream task of MT. Higher quality punctuation leads to translation BLEU gains for all seven target languages, as summarized in Table 5. The ST-Transformer system achieves the best results across all seven target languages. On average, the ST-Transformer system has a BLEU score improvement of 0.66 over the BL-Transformer and wins for all target languages. Similarly, the ST-LSTM system has a BLEU score improvement of 0.33 over the BL-LSTM system and wins for 5 out of 7 target languages. These results support our hypothesis.

Table 3. Punctuation results

| Test Set | Model | PERIOD | | | Q-MARK | | | COMMA | | | OVERALL | | | $F_1$-Gain |
|---|---|---|---|---|---|---|---|---|---|---|---|---|---|---|
| | | P | R | $F_1$ | P | R | $F_1$ | P | R | $F_1$ | P | R | $F_1$ | |
| *Dict-100* | BL-LSTM | 64 | 71 | 67 | 47 | 88 | 61 | 62 | 52 | 57 | 63 | 61 | 61 | |
| | ST-LSTM | 77 | 63 | 69 | 67 | 71 | 69 | 60 | 52 | 56 | 68 | 57 | 62 | 0.6% |
| | BL-Transformer | 69 | 76 | 72 | 50 | 88 | 64 | 68 | 52 | 59 | 68 | 63 | 65 | |
| | ST-Transformer | 81 | 71 | 76 | 82 | 82 | 82 | 69 | 51 | 59 | 74 | 60 | 67 | 2.9% |
| *MAEC* | BL-LSTM | 68 | 79 | 73 | 46 | 44 | 45 | 63 | 50 | 56 | 65 | 63 | 64 | |
| | ST-LSTM | 77 | 70 | 73 | 65 | 45 | 54 | 60 | 51 | 55 | 68 | 60 | 64 | 0.0% |
| | BL-Transformer | 71 | 80 | 75 | 50 | 50 | 50 | 65 | 49 | 56 | 67 | 63 | 65 | |
| | ST-Transformer | 80 | 78 | 79 | 69 | 46 | 56 | 65 | 48 | 55 | 72 | 62 | 66 | 2.4% |
| *EP-100* | BL-LSTM | 56 | 71 | 63 | 64 | 62 | 63 | 55 | 47 | 51 | 56 | 58 | 56 | |
| | ST-LSTM | 70 | 62 | 66 | 69 | 55 | 61 | 57 | 49 | 53 | 63 | 55 | 59 | 4.2% |
| | BL-Transformer | 58 | 76 | 66 | 58 | 70 | 64 | 57 | 49 | 53 | 57 | 61 | 59 | |
| | ST-Transformer | 70 | 71 | 71 | 76 | 70 | 73 | 59 | 51 | 55 | 64 | 60 | 62 | 5.8% |
| *NPR-76* | BL-LSTM | 72 | 71 | 72 | 71 | 66 | 69 | 65 | 58 | 61 | 69 | 65 | 67 | |
| | ST-LSTM | 82 | 71 | 76 | 76 | 69 | 73 | 65 | 59 | 62 | 74 | 66 | 70 | 4.0% |
| | BL-Transformer | 76 | 77 | 76 | 76 | 70 | 73 | 68 | 60 | 64 | 72 | 69 | 71 | |
| | ST-Transformer | 87 | 79 | 83 | 81 | 75 | 78 | 70 | 61 | 65 | 79 | 71 | 75 | 6.0% |

Table 4. Segmentation results

| Test Set | Model | Segmentation | | | | | |
|---|---|---|---|---|---|---|---|
| | | P | R | $F_1$ | $F_1$-gain | $F_{0.5}$ | $F_{0.5}$-gain |
| *Dict-100* | BL-LSTM | 62 | 68 | 65 | | 63 | |
| | ST-LSTM | 74 | 60 | 66 | 1.5% | 71 | 12.0% |
| | BL-Transformer | 66 | 74 | 70 | | 67 | |
| | ST-Transformer | 79 | 69 | 73 | 4.3% | 77 | 13.8% |
| *MAEC* | BL-LSTM | 66 | 76 | 71 | | 68 | |
| | ST-LSTM | 76 | 68 | 72 | 1.4% | 74 | 9.5% |
| | BL-Transformer | 69 | 77 | 73 | | 70 | |
| | ST-Transformer | 79 | 75 | 77 | 5.5% | 78 | 10.9% |
| *EP-100* | BL-LSTM | 53 | 67 | 59 | | 55 | |
| | ST-LSTM | 66 | 58 | 62 | 5.1% | 64 | 16.1% |
| | BL-Transformer | 54 | 72 | 62 | | 57 | |
| | ST-Transformer | 67 | 68 | 68 | 9.7% | 67 | 18.2% |
| *NPR-76* | BL-LSTM | 71 | 70 | 70 | | 71 | |
| | ST-LSTM | 81 | 70 | 75 | 7.1% | 79 | 10.9% |
| | BL-Transformer | 74 | 75 | 75 | | 74 | |
| | ST-Transformer | 85 | 79 | 81 | 8.0% | 84 | 12.5% |

## 6.3. Downstream Task: Machine Translation

We measure the impact of segmentation and punctuation improvements on the downstream task of MT. Higher quality punctuation leads to translation BLEU gains for all 7 target languages, as summarized in Table 5. The ST-Transformer system achieves the best results across all 7 target languages. On average, the ST-Transformer system has a BLEU gain of 0.66 over BL-Transformer and wins for all target languages. Similarly, the ST-LSTM system has a BLEU gain of 0.33 over BL-LSTM system and wins for 5 out of 7 target languages. These results support our hypothesis.

We used Azure Cognitive Services Translator API and compared them with reference translations. For Portuguese (pt) and French (fr), ST-LSTM regresses slightly, while ST-Transformer outperforms BL-Transformer. It is worth noting that ST-Transformer achieves significant gains over BL-Transformer, +1.1 for German (de) and +1.4 for Greek (el). The results suggest that punctuation has a higher impact on translation accuracy for some language pairs. For some language pairs, translation is more robust to punctuation errors.

Table 5. Translation BLEU Results: English audio recognized, punctuated, and translated to 7 languages

| Language | Model | BLEU | Gain |
|---|---|---|---|
| de | BL-LSTM | 36.0 | |
| | ST-LSTM | 36.6 | +0.6 |
| | BL-Transformer | 36.4 | |
| | ST-Transformer | 37.5 | +1.1 |
| el | BL-LSTM | 39.8 | |
| | ST-LSTM | 40.8 | +1.0 |
| | BL-Transformer | 40.3 | |
| | ST-Transformer | 41.7 | +1.4 |
| fr | BL-LSTM | 41.0 | |
| | ST-LSTM | 40.6 | -0.4 |
| | BL-Transformer | 41.7 | |
| | ST-Transformer | 41.8 | +0.1 |
| It | BL-LSTM | 35.2 | |
| | ST-LSTM | 35.5 | +0.3 |
| | BL-Transformer | 35.4 | |
| | ST-Transformer | 35.9 | +0.5 |
| pl | BL-LSTM | 30.2 | |
| | ST-LSTM | 30.9 | +0.7 |
| | BL-Transformer | 31.1 | |
| | ST-Transformer | 31.7 | +0.6 |
| pt | BL-LSTM | 33.2 | |
| | ST-LSTM | 33 | -0.2 |
| | BL-Transformer | 33.7 | |
| | ST-Transformer | 33.9 | +0.2 |
| ro | BL-LSTM | 39.8 | |
| | ST-LSTM | 40.1 | +0.3 |
| | BL-Transformer | 40.5 | |
| | ST-Transformer | 41.2 | +0.7 |

Table 6. BL-Transformer's incorrect punctuation leads to incorrect translations from English to several languages. ST-Transformer correctly punctuates, resulting in correct translations.

| Language | BL-Transformer | ST-Transformer |
|---|---|---|
| en | I. Just have to share the view ... | I just have to share the view ... |
| de | I. Ich muss nur die Ansicht teilen ... | Ich muss nur die Ansicht teilen ... |
| fr | I. Il suffit de partager le point de ... | Je dois simplement partager le point de ... |
| it | I. Basti condividere l'opinione ... | Devo solo condividere l'opinione ... |

Table 6 presents an example of how incorrect punctuation can lead to downstream consequences in translated outputs. Here BL-Transformer incorrectly punctuates after "I" which results in (1) failure to translate the word, (2) incorrect translations for the subsequent text, and (3) incorrect punctuation in the translations to all languages. ST-Transformer, however, correctly punctuates and thus produces correct translations. This example demonstrates the importance of punctuation quality for downstream tasks such as MT.

## 7. CONCLUSION

Long pauses and hesitations occur naturally in dictation. We started this work to solve the over-segmentation problem for long-form dictation users. We discovered these elements affect other long-form transcription scenarios like conversations, meeting transcriptions, and broadcasts. Our streaming punctuation approach improves punctuation for a variety of these ASR scenarios. Higher quality punctuation directly leads to higher quality downstream tasks, for instance, improvement in BLEU scores in MT. We also established the effectiveness of streaming punctuation across the transformer and LSTM tagging models, thus establishing the robustness of streaming punctuation to different model architectures.

In this paper, we focused on improving punctuation for hybrid ASR systems. Our preliminary analysis has found that though end-to-end (E2E) ASR systems produce better punctuation out-of-the-box, they still don't fully solve over-segmentation and could benefit from streaming re-punctuation techniques. We plan to present our findings in the future. Streaming punctuation discussed here relies primarily on linguistic features and discards acoustic signals. We plan to extend this work with prosody-aware punctuation models. As we explore our method's effectiveness and potential for other languages, we are also interested in exploring the impact of intonation or accents on streaming punctuation.